\documentclass[conference]{IEEEtran}
\IEEEoverridecommandlockouts
\usepackage{cite}
\usepackage{tabularx,booktabs}
\usepackage{amsmath,amssymb,amsfonts}
\usepackage{algorithmic}
\usepackage{graphicx}
\usepackage{textcomp}
\usepackage{xcolor}
\usepackage[]{footmisc}
\newcolumntype{C}{>{\centering\arraybackslash}X}
\usepackage{lipsum}
\def\BibTeX{{\rm B\kern-.05em{\sc i\kern-.025em b}\kern-.08em
    T\kern-.1667em\lower.7ex\hbox{E}\kern-.125emX}}
\begin{document}

\title{A Deep Ensemble Framework for Fake News Detection and Classification \\
{\footnotesize \textsuperscript{*}Fake News Detection using Deep Learning }
}
\author{Arjun Roy, Kingshuk Basak, Asif Ekbal, Pushpak Bhattacharyya \\
  Department of 
  Computer Science 
  and Engineering \\
  Indian Institute of Technology Patna, India\\
  {\tt (arjun.mtmc17, kingshuk.mtcs16, asif, pb) @iitp.ac.in}\\} 


\maketitle

\begin{abstract}
Fake news, rumor, incorrect information, and misinformation detection are nowadays crucial issues as these might have serious consequences for our social fabrics. The rate of such information is increasing rapidly due to the availability of enormous web information sources including social media feeds, news blogs, online newspapers etc.  
 In this paper, we develop various deep learning models for detecting fake news and classifying them into the pre-defined fine-grained categories. 
 At first, we develop models based on Convolutional Neural Network (CNN) and Bi-directional Long Short Term Memory (Bi-LSTM) networks. The representations obtained from these two models are fed into a Multi-layer Perceptron Model (MLP) for the final classification. Our experiments on a benchmark dataset show promising results with an overall accuracy of 44.87\%, which outperforms the current state of the art.  
\end{abstract}


\section{Introduction}
''We live in a time of fake news- things that are made up and manufactured.''-by Neil Portnow.\\
Fake news, rumors, incorrect information, misinformation have grown tremendously due to the phenomenal growth in web information. During the last few years, there has been a year-on-year growth in information emerging from various social media networks, blogs, twitter, facebook etc. Detecting fake news, rumor in proper time is very important as otherwise, it might cause damage to social fabrics.  This has gained a lot of interest worldwide due to its impact on recent politics and its negative effects. In fact, Fake News has been named as 2017's word of the year by Collins dictionary{\footnote{ \label{f1}
http://www.thehindu.com/books/fake-news-named-word-of-the-year-2017/article19969519.ece}}. 

Many recent studies have claimed that US election 2016 was heavily impacted by the spread of Fake News.  
False news stories have become a part of everyday life, exacerbating weather crises, political violence, intolerance between people of different ethnics and culture, and even affecting matters of public health.  
All the governments around the world are trying to track and address these problems. 
On 1$^{st}$ Jan, 2018, bbc.com published that ''Germany is set to start enforcing a law that demands social media sites move quickly to remove hate speech, fake news, and illegal material.''
Thus it is very evident that the development of automated techniques for detection of Fake News is very important and urgent.
\subsection{Problem Definition and Motivation}
Fake News can be defined as completely misleading or made up information that is being intentionally circulated claiming as true information. In this paper, we develop a deep learning based system for detecting fake news.  

Deception detection is a well-studied problem in Natural Language Processing (NLP) and researchers have addressed this problem quite extensively. 
The problem of detecting fake news in our everyday life, although very much related to deception detection, but in practice is much more challenging and hard, as the news body often contains a very few and short statements. Even for a human reader, it is difficult to accurately distinguish true from false information by just looking at these short pieces of information. Developing suitable hand engineered features (for a classical supervised machine learning model) to identify fakeness of such statements is also a technically challenging task. In contrast to classical feature-based model, deep learning has the advantage in the sense that it does not require any handcrafting of rules and/or features, rather it identifies the best feature set on its own for a specific problem. 
For a given news statement, our proposed technique classifies the short statement into the following fine-grained classes: \textit{true}, \textit{mostly-true}, \textit{half-true}, \textit{barely-true}, \textit{false} and \textit{pants-fire}. Example of an instance of each class is given in Table \ref{example}. 
\begin{center}
\begin{table*}[ht!]
\resizebox{\textwidth}{!}{
\begin{tabular}{|c|c|c|c|c|c|c|c|c|c|c|c|c|c|}
\hline
\textbf{Label} & \textbf{Statement}                                                                                                                                & \textbf{\begin{tabular}[c]{@{}c@{}}Statement\\  Type\end{tabular}} & \textbf{Speaker} & \textbf{\begin{tabular}[c]{@{}c@{}}Speaker's\\  Job\end{tabular}} & \textbf{State} & \textbf{Party} & \textbf{Pants-fire} & \textbf{False} & \textbf{\begin{tabular}[c]{@{}c@{}}Barely\\ True\end{tabular}} & \textbf{\begin{tabular}[c]{@{}c@{}}Half\\  True\end{tabular}} & \textbf{\begin{tabular}[c]{@{}c@{}}Mostly\\ True\end{tabular}}& \textbf{Context}  \\ \hline

TRUE	& \begin{tabular}[c]{@{}c@{}}\\ McCain opposed a requirement\\ that the government \\buy American-made \\motorcycles. And he said \\all buy-American provisions \\were quote 'disgraceful.' \\ \end{tabular}	& federal-budget	& barack-obama	& President	& Illinois	& democrat	& 70	& 71	& 160	& 163	& 9	& a radio ad	
\\ \hline
mostly-true &	\begin{tabular}[c]{@{}c@{}}\\Almost 100,000 people left \\Puerto Rico last year.\end{tabular}	& \begin{tabular}[c]{@{}c@{}} bankruptcy,\\economy,\\population \end{tabular}	& jack-lew & \begin{tabular}[c]{@{}c@{}}	Treasury \\secretary \end{tabular} &	Washington, D.C. & 	democrat	& 0 &	1 &	0	& 1 &	0	& \begin{tabular}[c]{@{}c@{}} an interview \\ with Bloomberg \\News \end{tabular}
\\ \hline
half-true	& \begin{tabular}[c]{@{}c@{}}\\ Rick Perry has never lost \\an election and \\remains the only person to have\\ won the Texas governorship three \\times in landslide elections.\\
\end{tabular} & candidates-biography
 &	ted-nugent
 & 	musician
	&	Texas
	& republican
	& 0	& 0	& 2	& 0	& 2	& \begin{tabular}[c]{@{}c@{}}an oped \\column.
\end{tabular} \\ \hline
barely-true	& \begin{tabular}[c]{@{}c@{}} \\Says Mitt Romney wants to get \\rid of Planned Parenthood.\\
\end{tabular} & \begin{tabular}[c]{@{}c@{}}abortion,\\federal-budget,\\health-care\end{tabular}
 & \begin{tabular}[c]{@{}c@{}}	planned-parenthood\\-action-fund \end{tabular}
 & 	Advocacy group
	&	Washington, D.C.
	& none
	& 1	& 0	& 0	& 0	& 0	& a radio ad \\ \hline
FALSE	& \begin{tabular}[c]{@{}c@{}} \\I dont know\\ who (Jonathan Gruber) is.\\
\end{tabular}  &	health-care & 	nancy-pelosi

 & 	\begin{tabular}[c]{@{}c@{}} 	House Minority \\Leader \end{tabular} 
	&	California
	& democrat
	& 3	& 7	& 11	& 2	& 3	& \begin{tabular}[c]{@{}c@{}} a news\\ conference \end{tabular}\\ \hline
   pants-fire	& \begin{tabular}[c]{@{}c@{}} \\Transgender individuals in the U.S.\\ have a 1-in-12 chance of \\being murdered.
\\
\end{tabular}  &	\begin{tabular}[c]{@{}c@{}}corrections-and-updates,\\crime,criminal-justice,\\sexuality\end{tabular}
 & 	garnet-coleman

 & 	\begin{tabular}[c]{@{}c@{}} 	president, \\ceo of Apartments \\for America, Inc.
\end{tabular} 
	&	Texas

	& democrat
	& 1	& 0	& 1	& 0	& 1	& \begin{tabular}[c]{@{}c@{}} a committee \\hearing
 \end{tabular}\\ \hline
\end{tabular}
}
\caption{Example of instances of each class.}
\label{example}
\end{table*}
\end{center}
\subsection{Contributions}
Most of the existing studies on fake news detection are based on classical supervised model. In recent times there has been an interest towards developing deep learning based fake news detection system, but these are mostly concerned with binary classification.
In this paper, we attempt to develop an ensemble based architecture for fake news detection. The individual models are based on Convolutional Neural Network (CNN) and Bi-directional Long Short Term Memory (LSTM). The representations obtained from these two models are fed into a Multi-layer Perceptron (MLP) for multi-class classification.
\subsection{Related Work}\label{related}
The concept of fake news is often linked with rumor, deception, hoax, spam etc. Some of the related work can be found in \cite{7} for rumour, \cite{10} for deception detection, \cite{15} for hoax, and \cite{16} for spam. 
Problems related to these topics have mostly been viewed with respect to binary classification. Likewise, most of the published works also has viewed fake news detection as a binary classification problem (i.e fake or true). Bajaj\cite{2} in his work applied various deep learning strategies on dataset composed of fake news articles available in Kaggle\footnote{https://www.kaggle.com/mrisdal/fake-news} and authentic news articles extracted from Signal Media News\footnote{http://research.signalmedia.co/newsir16/signal-dataset.html} dataset and observed that classifiers based on Gated Recurrent Unit (GRU), Long Short Term Memory (LSTM), Bi-directional Long Short Term Memory (Bi-LSTM) performed better than the classifiers based on CNN. Natali Ruchansky et al.\cite{1} used social media dataset (which is also used in \cite{6} for Rumor Detection) and developed a hybrid deep learning model which showed the accuracies of 0.892 on Twitter data and 0.953 on Weibo data. They showed that both, capturing the temporal behavior of the articles as well as learning source characteristics about the behavior of the users, are important for fake news detection. Further integrating these two elements improves the performance of the classifier. 

By observing very closely it can be seen that fake news articles can be classified into multiple classes depending on the fakeness of the news. For instance, there can be certain exaggerated or misleading information attached to a true statement or news. Thus, the entire news or statement can neither be accepted as completely true nor can be discarded as entirely false. This problem was addressed by William Y Yang in his paper \cite{8} where he introduced \textbf{Liar} dataset comprising of a substantial volume of short political statements having six different class annotations determining the amount of fake content of each statement. In his work, he showed comparative studies of several statistical and deep learning based models for the classification task and found that the CNN model performed best with an accuracy of 0.27. Y. Long et al. \cite{19} in their work used the \textbf{Liar} \cite{19} dataset, and proposed a hybrid attention-based LSTM model for this task, which outperformed W.Yang's hybrid CNN model by 14.5\% in accuracy, establishing a new state of the art.

In our current work we propose an ensemble architecture based on CNN \cite{13} and Bi-LSTM \cite{11}, and this has been evaluated on \textbf{Liar}\cite{8} dataset. Our proposed model tries to capture the pattern of information from the short statements and learn the characteristic behavior of the source speaker from the different attributes provided in the dataset, and finally integrate all the knowledge learned to produce fine-grained multi-class classification.   
\section{Methodology }\label{methods}
We propose a deep multi-label classifier for classifying a statement into six fine-grained classes of fake news. Our approach is based on an ensemble model that makes use of Convolutional Neural Network (CNN) \cite{13} and Bi-directional Long Short Term Memory (Bi-LSTM) \cite{11}. The information presented in a statement is essentially sequential in nature. In order to capture such sequential information we use Bi-LSTM architecture. Bi-LSTM is known to capture information in both the directions: forward and backward. Identifying good features manually to separate true from fake even for binary classification, is itself, a technically complex task as human expert even finds it difficult to differentiate true from the fake news. Convolutional Neural Network (CNN) is known to  capture the hidden features efficiently. 
We hypothesize that CNN will be able to detect hidden features of the given statement and the information related to the statements to eventually judge the authenticity of each statement. We make an intuition that both- capturing temporal sequence and identifying hidden features, will be necessary to solve the problem. As described in data section, each short statement is associated with 11 attributes that depict different informations regarding the speaker and the statement. After our thorough study we identify the following relations among the various attributes which contribute towards labeling of the given statements.
\begin{enumerate}\label{relation}
\item Relation between \textbf{Statement} and \textbf{Statement type}
\item Relation between \textbf{Statement} and \textbf{Context}
\item Relation between \textbf{Speaker} and \textbf{Party}.
\item Relation between \textbf{Party} and \textbf{Speaker's job}.
\item Relation between \textbf{Statement type} and \textbf{Context}.
\item Relation between \textbf{Statement} and \textbf{State}.
\item Relation between \textbf{Statement} and \textbf{Party}.
\item Relation between \textbf{State} and \textbf{Party}.
\item Relation between \textbf{Context} and \textbf{Party}.
\item Relation between \textbf{Context} and \textbf{Speaker}.\\
\end{enumerate}
To ensure that deep networks understand these relations we propose to feed each of these relations into separate network layers and eventually after extracting all the relations we group them together along with the five-column attributes containing information regarding speaker's total credit history count. In addition to that, we also feed in a special feature vector that is proposed by us and is to be formed using the count history information. This vector is a five digit number signifying the five count history columns, with only one of the digit being set to '1' (depending on which column has the highest count) and the rest of the four digits are set to '0'. 
\subsection{Bi-LSTM}
Bidirectional LSTMs are the networks with LSTM units that process word sequences in both the directions (i.e. from left to right as well as from right to left). In our model we consider the maximum input length of each statement to be 50 (average length of statements is 17 and the maximum length is 66, and only 15 instances of the training data of length greater than 50) with post padding by zeros. For attributes like statement type, speaker's job, context we consider the maximum length of the input sequence to be 5, 20, 25, respectively. Each input sequence is embedded into 300-dimensional vectors using pre-trained Google News vectors\cite{12} (Google News Vectors 300dim is also used in \cite{8} for embedding). Each of the embedded inputs are then fed into separate Bi-LSTM networks, each having 50 neural units at each direction. The output of each of these Bi-LSTM network is then passed into a dense network of 128 neurons with activation function as 'ReLU'.   
\begin{figure}
\label{bilstm}
\end{figure}

\subsection{CNN}
Over the last few years many experimenters
has shown that the convolution and pooling
functions of CNN can be successfully used
to find out hidden features of not only images
but also texts. A convolution layer of \(n \times m\)
kernel size will be used (where m-size of word
embedding) to look at n-grams of words at a time and then a MaxPooling layer will select the largest
from the convoluted inputs, as shown in
Figure \ref{cnn}. 
\begin{figure}
\end{figure}
The attributes, namely speaker, party, state are embedded using pre-trained 300-dimensional Google News Vectors\cite{12} and then the embedded inputs are fed into separate Conv layers.
The different credit history counts the fake statements of a speaker and a feature proposed by us formed using the credit history counts are directly passed into separate Conv layers.

\subsection{Combined CNN and Bi-LSTM Model}
The representations obtained from CNN and Bi-LSTM are combined together to obtain better performance. 

The individual dense networks following the Bi-LSTM networks carrying information about the statement, the speaker's job, context are reshaped and then passed into different Conv layers. Each convolution layer is followed by a Maxpooling layer, which is then flattened and passed into separate dense layers.
Each of the dense layers of different networks carrying different attribute information are merged, two at a time-to capture the relations among the various attributes as mentioned at the beginning of \ref{relation}. Finally, all the individual networks are merged together and are passed through a dense layer of six neurons with softmax as activation function as depicted in. The classifier is optimized using Adadelta as optimization technique with categorical cross-entropy as the loss function. 

\begin{figure}
\includegraphics[width=0.4\textwidth]{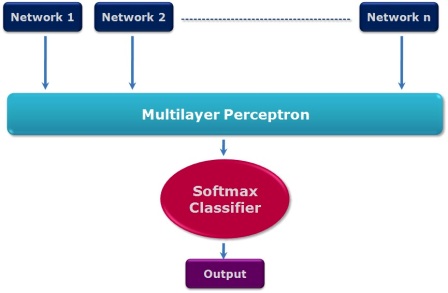}
\caption{Ensemble based architecture } \label{ensem}
\end{figure}
\section{Data}\label{data}
We use the dataset, named \textbf{LIAR} (Wang 2017), for our experiments. The dataset is annotated with six fine-grained classes and comprises of about 12.8K annotated short statements along with various information about the speaker. The statements which were mostly reported during the time interval [2007 to 2016], are considered for labeling by the editors of \textbf{Politifact.com}. Each row of the data contains a short statement, a label of the statement and 11 other columns correspond to various information about the speaker of the statement. Descriptions of these attributes are given below: 
\begin{enumerate}
\item \textbf{Label}: Each row of data is classified into six different types, namely
\begin{enumerate}
\item \textbf{Pants-fire:} Means the speaker has delivered a blatant lie .
\item \textbf{False:} Means the speaker has given totally false information.
\item \textbf{Barely-true:} Chances of the statement depending on the context is hardly true. Most of the contents in the statements are false.
\item \textbf{Half-true:} Chances of the content in the statement is approximately half. 
\item \textbf{Mostly-true:} Most of the contents in the statement are true.
\item \textbf{True:} Content is true.
\end{enumerate}
\item \textbf{Statement by the politician:} This statement is a short statement.
\item \textbf{Subjects:} This corresponds to the content of the text. For examples, foreign policy, education, elections etc.
\item \textbf{Speaker:} This contains the name of the speaker of the statement.
\item \textbf{Speaker's job title:} This specifies the position of the speaker in the party.
\item \textbf{State information:} This specifies in which state the statement was delivered.
\item \textbf{Party affiliation:} This denotes the name of the party of the speaker belongs to.
\item The next five columns are the counts of the speaker's statement history. They are:
\begin{enumerate}
\item \textbf{Pants fire count;}
\item \textbf{False count;}
\item \textbf{Barely true count;}
\item \textbf{Half false count;}
\item \textbf{Mostly true count.}
\end{enumerate}
\item \textbf{Context:} This corresponds to the venue or location of the speech or statement.\\
\end{enumerate}
The dataset consists of three sets, namely a training set of 10,269 statements, a validation set of 1,284 statements and a test set of 1,266 statements. 
\section{Experiments and Results }\label{res}
In this section, we report on the experimental setup, evaluation results, and the necessary analysis. 
\subsection{Experimental Setup}
All the experiments are conducted in a python environment.
The libraries of python are required for carrying out the experiments are \textbf{Keras}, \textbf{NLTK}, \textbf{Numpy}, \textbf{Pandas}, \textbf{Sklearn}.
We evaluate the performance of the system in terms of accuracy, precision, recall, and F-score metrics. 

\subsection{Results and Analysis}
We report the evaluation results in Table \ref{results} that also show the comparison with the system as proposed in \cite{8} and \cite{19}. 
\begin{table}[h!]

\resizebox{\columnwidth}{!}{%
\begin{tabular}{|c|c|c|c|}
\hline
\textbf{Model}               & \textbf{Network} & \textbf{Attributes taken} & \textbf{Accuracy} \\ \hline
\textbf{William Yang Wang\cite{8}} & Hybrid CNN     & All                       & 0.274            \\ 
\hline
\textbf{Y. Long et al.\cite{19}} & Hybrid LSTM     & All                       & 0.415            \\ \hline
\textbf{Bi-LSTM Model}          & Bi-LSTM        & All                       & 0.4265             \\ \hline
\textbf{CNN Model}          & CNN        & All                       & 0.4289             \\ \hline
\textbf{Our Proposed Model}          & RNN-CNN combined        & All                       & \textbf{0.4487}             \\ \hline
\end{tabular}
}
\centering
\caption{Overall evaluation results }
\label{results}
\end{table}
We depict the overall evaluation results in Table \ref{results} along with the other existing models. This shows that our model performs better than the existing state-of-the-art model as proposed in \cite{19}. This state-of-the-art model was a hybrid LSTM, with an accuracy of 0.415. 
On the other hand, our proposed model shows 0.4265, 0.4289 and 0.4487 accuracies for Bi-LSTM, CNN and the combined CNN+Bi-LSTM model, respectively.  
This clearly supports our assumption that capturing temporal patterns using Bi-LSTM and hidden features using CNN are useful, channelizing each profile attribute through a different neural layer is important, and the meaningful combination of these separate attribute layers to capture relations between attributes, is effective.
\begin{center}
\begin{table}
\resizebox{\columnwidth}{!}{
\begin{tabular}{l|l|l|l|l|}

\cline{2-5}
                                           & \textbf{precision} & \textbf{recall} & \textbf{F1-score} & \textbf{No. of instances} \\ \hline
\multicolumn{1}{|l|}{\textbf{PANTS-FIRE}}  & 0.73      &0.35      &0.47               & 92               \\ \hline
\multicolumn{1}{|l|}{\textbf{FALSE}}       & 0.47      &0.53    &  0.50             & 249              \\ \hline
\multicolumn{1}{|l|}{\textbf{BARELY-TRUE}} & 0.58     & 0.32    &  0.41              & 212              \\ \hline
\multicolumn{1}{|l|}{\textbf{HALF-TRUE}}  & 0.39     & 0.46      &0.42              & 265              \\ \hline
\multicolumn{1}{|l|}{\textbf{MOSTLY-TRUE}} & 0.33    &  0.66      &0.44             & 241             \\ \hline
\multicolumn{1}{|l|}{\textbf{TRUE}}        &  0.88     & 0.14     & 0.23              & 207              \\ \hline
\multicolumn{1}{|l|}{\textbf{Avg/Total}}   & 0.53     & 0.43     & 0.41             & 1266             \\ \hline
\end{tabular}
}
\caption{Evaluation of Bi-LSTM model: precision, recall, and F1 score }
\label{preci-bi}
\end{table}

\begin{table}
\resizebox{\columnwidth}{!}{%
\begin{tabular}{l|l|l|l|l|}

\cline{2-5}
                                           & \textbf{precision} & \textbf{recall} & \textbf{F1-score} & \textbf{No. of instances } \\ \hline
\multicolumn{1}{|l|}{\textbf{PANTS-FIRE}}  &  0.67     & 0.39     & 0.49               & 92               \\ \hline
\multicolumn{1}{|l|}{\textbf{FALSE}}       & 0.36     & 0.63     & 0.46              & 249              \\ \hline
\multicolumn{1}{|l|}{\textbf{BARELY-TRUE}} & 0.50    &  0.36     & 0.42              & 212              \\ \hline
\multicolumn{1}{|l|}{\textbf{HALF-TRUE}}   & 0.42     & 0.46    &  0.44              & 265              \\ \hline
\multicolumn{1}{|l|}{\textbf{MOSTLY-TRUE}} & 0.41     & 0.49      &0.45             & 241             \\ \hline
\multicolumn{1}{|l|}{\textbf{TRUE}}        & 0.70     & 0.16      &0.26             & 207              \\ \hline
\multicolumn{1}{|l|}{\textbf{Avg/Total}}   &  0.48      &0.43     & 0.42              & 1266             \\ \hline
\end{tabular}
}
\caption{Evaluation of CNN model: precision, recall, F1 score}
\label{preci-cnn}
\end{table}

\begin{table*}[h!]
\resizebox{\textwidth}{!}{
\begin{tabular}{|c|c|c|c|c|c|c|c|c|c|c|c|c|c|}
\hline
\textbf{Label} & \textbf{Statement}                                                                                                                                & \textbf{\begin{tabular}[c]{@{}c@{}}Statement\\  Type\end{tabular}} & \textbf{Speaker} & \textbf{\begin{tabular}[c]{@{}c@{}}Speaker's\\  Job\end{tabular}} & \textbf{State} & \textbf{Party} & \textbf{Context} & \textbf{Pants-fire} & \textbf{False} & \textbf{\begin{tabular}[c]{@{}c@{}}Barely\\ True\end{tabular}} & \textbf{\begin{tabular}[c]{@{}c@{}}Half\\  True\end{tabular}} & \textbf{\begin{tabular}[c]{@{}c@{}}Mostly\\ True\end{tabular}} & \textbf{\begin{tabular}[c]{@{}c@{}}Predicted \\ Label\end{tabular}} \\ \hline
barely-true    & \begin{tabular}[c]{@{}c@{}}We know there are\\more  Democrats in \\ Georgia than Republicans.\\  We know that for a fact.\end{tabular} & elections                                                          & mike-berlon      & none                                                              & Georgia        & democrat       & an article       & 1                   & 0              & 0                                                              & 0                                                             & 0                                                              & False                                                               \\ \hline
\end{tabular}
}
\caption{Sample text with wrongly predicted label and original label.}
\label{error predict}
\end{table*}
\end{center}
We also report the precision, recall and F-score measures for all the models. Table\ref{preci-bi}, Table \ref{preci-cnn} and Table \ref{preci-comb} depict the evaluation results of CNN, Bi-LSTM and the combined model of CNN and Bi-LSTM, respectively.
The evaluation shows that on the precision measure the combined model performs best with an average precision of \textbf{0.55} while that of Bi-LSTM model is 0.53 and CNN model is 0.48. The combined model of CNN and Bi-LSTM even performs better with respect to recall and F1-Score measures. The combined model yields the average recall of \textbf{0.45} and average F1-score of \textbf{0.43} while that of Bi-LSTM model is 0.43 and 0.41, respectively and of the CNN model is 0.43 and 0.42, respectively. On further analysis, we observe that although the performance (based on precision, recall, and F1-score) of each of the models for every individual class is close to the average performance, but in case of the prediction of the class label \textbf{TRUE} the performance of each model varies a lot from the respective average value. The precisions of TRUE is promising (Bi-LSTM model:0.88, CNN model: 0.7, Combined model:\textbf{0.85}), but the recall (Bi-LSTM model:0.14, CNN model: 0.16, Combined model:0.14) and the F1-score (Bi-LSTM model:0.23, CNN model: 0.26, Combined model:0.24) are very poor. This entails the fact that our proposed model predicts comparatively less number of instances as TRUE, but when it does the prediction is very accurate. Thus it can be claimed that if a statement is predicted as \textbf{True} by our proposed model then one can rely on that with high confidence.
Although our model performs superior compared to the existing state-of-the-art, still the results were not error free. We closely analyze the models' outputs to understand their behavior and perform both quantitative as well as qualitative error analysis. For quantitative analysis, we create the confusion matrix for each of our models. Confusion matrix corresponding to the experiment 1 i.e with Bi-LSTM model is given in Table\ref{confu-bi}, corresponding  to experiment 2 i.e with CNN model is given in Table\ref{confu-cnn} and corresponding to our final experiment i.e with RNN-CNN combined model is given in Table \ref{confu-comb}.\\~\\
From these quantitative analysis it is seen that in majority of the cases the test data statements originally labeled with \textbf{Pants-Fire} class gets confused with the \textbf{False} class, statements originally labeled as \textbf{False} gets confused with \textbf{Barely true} and \textbf{half true} classes, statements originally labeled as \textbf{Half true} gets confused with \textbf{Mostly True} and \textbf{False} class, statements originally labeled as \textbf{Mostly true} gets confused with \textbf{Half True}, statements originally labeled with \textbf{True} gets confused with \textbf{Mostly True} class.\\~\\
It is quite clear that errors were mostly concerned with the classes, overlapping in nature. Confusion is caused as the contents of the statements belonging to these classes are quite similar. For example, the difference between 'Pants-Fire' and 'False' class is that only the former class corresponds to the false information with more intensity. Likewise 'Half True' has high similarity to 'False', and 'True' with 'Mostly True'. The difference between `True' and `Mostly True' is that the later class has some marginal amount of false information, while the former does not. 

\begin{center}
\begin{table}
\resizebox{\columnwidth}{!}{%

\begin{tabular}{l|l|l|l|l|}

\cline{2-5}
                                           & \textbf{precision } & \textbf{recall} & \textbf{F1-score} & \textbf{No. of instances } \\ \hline
\multicolumn{1}{|l|}{\textbf{PANTS-FIRE}}  & 0.70    &  0.43    &  0.54               & 92               \\ \hline
\multicolumn{1}{|l|}{\textbf{FALSE}}       & 0.45      &0.61     & 0.52              & 249              \\ \hline
\multicolumn{1}{|l|}{\textbf{BARELY-TRUE}} & 0.61    &  0.32     & 0.42             & 212              \\ \hline
\multicolumn{1}{|l|}{\textbf{HALF-TRUE}}   & 0.35     & 0.73     & 0.47             & 265              \\ \hline
\multicolumn{1}{|l|}{\textbf{MOSTLY-TRUE}} &0.50     & 0.36      &0.42             & 241             \\ \hline
\multicolumn{1}{|l|}{\textbf{TRUE}}        & 0.85     & 0.14     & 0.24              & 207              \\ \hline
\multicolumn{1}{|l|}{\textbf{Avg/Total}}   &\textbf{ 0.55}               & \textbf{0.45}            & \textbf{0.43}              & 1266             \\ \hline
\end{tabular}
}
\caption{Evaluation of Bi-LSTM, CNN combined model: precision, recall, F1 score }
\label{preci-comb}
\end{table}
\end{center}
For qualitative analysis, we closely look at the actual statements and try to understand the causes of misclassifications. We come up with some interesting facts.
There are some speakers whose statements are not present in the training set, but are present in the test set. For few of these statements, our model tends to produce wrong answers. Let us consider the example given in Table\ref{error predict}. For this speaker, there is no training data available and also the count history of the speaker is very less. So our models assign an incorrect class. But it is to be noted that even if there is no information about the speaker in the training data and the count history of the speaker is almost empty, still we are able to generate a prediction of a class that is close to the original class in terms of meaning. 

It is also true that classifiers often make mistakes in making the fine distinction between the classes due to the insufficient number of training instances. Thus, classifiers tend to misclassify the instances into one of the nearby (and overlapped) classes. 
\begin{table}[h!]
\centering
\resizebox{\columnwidth}{!}{
\begin{tabular}{|c|c|c|c|c|c|c|}
\hline
\textit{\textbf{Actual$\backslash$Predicted}} & \textbf{Pants-Fire} & \textbf{False} & \textbf{Barely-True} & \textbf{Half-True} & \textbf{Mostly-True} & \textbf{True} \\ \hline
\textbf{Pants-Fire}                & 32  &35  & 3  & 8  &14  & 0            \\ \hline
\textbf{False}                     & 4 &131 & 16 & 36 & 59 &  3             \\ \hline
\textbf{Barely-True}               & 5  &31 & 68 & 48 & 60 &  0             \\ \hline
\textbf{Half-True}                 &0  &38  & 8 &123 & 95 &  1             \\ \hline
\textbf{Mostly-True}               & 1  &20  & 8  &54 &158  & 0            \\ \hline
\textbf{True}                      & 2 & 25 & 15 & 47 & 90 & 28            \\ \hline
\end{tabular}
}
\caption{Confusion matrix of the Bi-LSTM model}
\label{confu-bi}
\end{table}

\begin{table}[h!]
\centering
\resizebox{\columnwidth}{!}{
\begin{tabular}{|c|c|c|c|c|c|c|}
\hline
\textit{\textbf{Actual$\backslash$Predicted}} & \textbf{Pants-Fire} & \textbf{False} & \textbf{Barely-True} & \textbf{Half-True} & \textbf{Mostly-True} & \textbf{True} \\ \hline
\textbf{Pants-Fire}                & 36 & 35 &  6 & 11 &  2 &  2           \\ \hline
\textbf{False}                     & 7 &156 & 21 & 30 & 28  & 7             \\ \hline
\textbf{Barely-True}               & 5  &66 & 76  &34  &29  & 2            \\ \hline
\textbf{Half-True}                 & 2  &75 & 14 &123 & 48  & 3             \\ \hline
\textbf{Mostly-True}               & 1 & 53  &17  &51 &119  & 0            \\ \hline
\textbf{True}                      &  3  &44  &18 & 44 & 65 & 33           \\ \hline
\end{tabular}
}
\caption{Confusion matrix of the CNN model}
\label{confu-cnn}
\end{table}

\begin{table}[h!]
\centering
\resizebox{\columnwidth}{!}{
\begin{tabular}{|c|c|c|c|c|c|c|}
\hline
\textit{\textbf{Actual$\backslash$Predicted}} & \textbf{Pants-Fire} & \textbf{False} & \textbf{Barely-True} & \textbf{Half-True} & \textbf{Mostly-True} & \textbf{True} \\ \hline
\textbf{Pants-Fire}                & 40 & 34  & 4 & 10  & 4  & 0             \\ \hline
\textbf{False}                     & 7 &152  &10  &67  &11   &2            \\ \hline
\textbf{Barely-True}               &4 & 48 & 68 & 83  & 9  & 0            \\ \hline
\textbf{Half-True}                 & 0 & 43  & 7 &193  &20  & 2            \\ \hline
\textbf{Mostly-True}               &  2 & 31 &  9 &112 & 86  & 1            \\ \hline
\textbf{True}                      & 4 & 31  &13 & 89  &41  &29           \\ \hline
\end{tabular}
}
\caption{Confusion matrix of the Bi-LSTM+CNN combined model}
\label{confu-comb}
\end{table}

\section{Conclusion and Future Works}
In this paper, we have tried to address the problem of fake News detection by looking into short political statements made by the speakers in different types of daily access media. The task was to classify any statement into one of the fine-grained classes of fakeness. We have built several deep learning models, based on CNN, Bi-LSTM and the combined CNN and Bi-LSTM model. 
 Our proposed approaches mainly differ from previously mentioned models in system architecture, and each model performs better than the state of the art as proposed in \cite{19}, where the statements were passed through one LSTM and all the other details about speaker's profile through another LSTM.
On the other hand, we have passed every different attribute of speaker's profile through a different layer, captured the relations between the different pairs of attributes by concatenating them. Thus, producing a meaningful vector representation of relations between speaker's attributes, with the help of which we obtain the overall accuracy of 44.87\%.
By further exploring the confusion matrices we found out that classes which are closely related in terms of meaning are getting overlapped during prediction. We have made a thorough analysis of the actual statements, and derive some interesting facts. There are some speakers whose statements are not present in the training set but present in the test set. For some of those statements, our model tends to produce the wrong answers. This shows the importance of speakers' profile information for the task. Also 
as the classes and the meaning of the classes are very near, they tend to overlap due to less number of examples in training data.

We would like like to highlight some of the possible solutions to solve the problems that we encountered while attempted to solve fake news detection problem in a more fine-grained way.
\begin{itemize}
\item More labeled data sets are needed to train the model more accurately. Some semi-supervised or active learning models might be useful for this task.
\item Along with the information of a speaker's count history of lies, the actual statements are also needed in order to get a better understanding of the patterns of the speaker's behavior while making a statement.

\end{itemize}

Fake news detection into finely grained classes that too from short statements is a challenging but interesting and practical problem. Hypothetically the problem can be related to \textbf{Sarcasm detection}\cite{14} problem. Thus it will also be interesting to see the effect of implementing the existing methods that are effective in sarcasm detection domain in Fake News detection domain.

\end{document}